# Binary Join Trees

**Prakash P. Shenoy**
School of Business
University of Kansas
Summerfield Hall
Lawrence, KS 66045-2003, USA
p-shenoy@ukans.edu
http://statl.cc.ukans.edu/~pshenoy

## Abstract

The main goal of this paper is to describe a data structure called binary join trees that are useful in computing multiple marginals efficiently using the Shenoy-Shafer architecture. We define binary join trees, describe their utility, and sketch a procedure for constructing them.

## 1 INTRODUCTION

The main goal of this paper is to describe a data structure called binary join trees that are useful in computing multiple marginals efficiently using the Shenoy-Shafer architecture. We define binary join trees, describe their utility, and sketch a procedure for constructing them.

In the last decade, much work has been done in the uncertain reasoning community on exact computation of marginals using local computation [see, e.g., Pearl 1986, Kong 1986, Lauritzen and Spiegelhalter 1988, Shenoy and Shafer 1990, Jensen et al. 1990]. The main idea behind local computation is to compute marginals of the joint distribution without actually computing the joint distribution. Local computation can be described as message passing in data structures called join trees. Join trees are also called junction trees [Jensen et al. 1990], clique trees [Lauritzen and Spiegelhalter 1988], qualitative Markov trees [Shafer et al. 1987], and hypertrees [Shenoy and Shafer 1990].

The efficiency of the message-passing algorithms depend on the sizes of the subsets in a join tree. The problem of finding a join tree that minimizes the size of the largest subset has been shown to be NP-complete [Arnborg et al. 1987]. Consequently, much attention has been devoted to finding heuristics for constructing good join trees [see, e.g., Olmsted 1983, Kong 1986, Mellouli 1987, Zhang 1988, Kjærulff 1990].

In this paper, we focus on another aspect of join trees, the number of neighbors of nodes in a join tree. If a node in a join tree has many neighbors, then it leads to much inefficiencies in the Shenoy-Shafer architecture. This motivates the definition of binary join trees which is a join tree such that no node has more than three neighbors.

The main idea behind a binary join tree is that all combinations are done on a binary basis, i.e., we combine functions two at a time.

Local computation has also been studied in many other domains besides uncertain reasoning such as solving systems of equations [Rose 1970], optimization [Bertele and Brioschi 1972], and relational databases [Beeri et al. 1983]. In order to keep the applicability of the results as wide as possible, we describe our work using the abstract framework of valuation networks [Shenoy 1989, 1992].

An outline of this paper is as follows. Section 2 introduces the valuation network framework. Section 3 describes the Shenoy-Shafer architecture for computing multiple marginals. Section 4 introduces the concept of binary join trees and its utility in reducing the computational effort of computing multiple marginals. Finally Section 5 contains concluding remarks.

## 2 THE VALUATION NETWORK FRAMEWORK

This section describes the abstract valuation network (VN) framework [Shenoy 1989, 1992]. In a VN, we represent knowledge by entities called valuations, and we make inferences using two operators called marginalization and combination that operate on valuations.

### 2.1 VARIABLES AND CONFIGURATIONS

We use the symbol $\Omega_X$ for the set of possible values of a variable X, and we call $\Omega_X$ the *state space* for X. We are concerned with a finite set $\Psi$ of variables, and we assume that all the variables in $\Psi$ have finite state spaces. We use upper-case Roman letters such as X, Y, Z, etc., to denote variables, and we use italicized lower-case Roman letters such as $r$, $s$, $t$, etc., to denote sets of variables.

Given a nonempty set $s$ of variables, let $\Omega_s$ denote the Cartesian product of $\Omega_X$ for $X \in s$; $\Omega_s = \times \{\Omega_X \mid X \in s\}$. We call $\Omega_s$ the *state space* for $s$. We call the elements of $\Omega_s$ *configurations* of $s$. We use lower-case, bold-faced letters such as $\mathbf{x}$, $\mathbf{y}$, $\mathbf{z}$, etc., to denote configurations.

## 2.2 VALUATIONS

Given a subset $s$ of variables (possibly empty), there is a set $\vartheta_s$. We call the elements of $\vartheta_s$ *valuations for s*. Let $\vartheta$ denote the set of all valuations, i.e., $\vartheta = \cup \{\vartheta_s \mid s \subseteq \Psi\}$. If $\sigma$ is a valuation for $s$, we say $s$ is the *domain* of $\sigma$. We use lower-case Greek letters such as $\rho$, $\sigma$, $\tau$, etc., to denote valuations.

## 2.3 MARGINALIZATION

We assume that for each nonempty $s \subseteq \Psi$, and for each X $\in s$, there is a mapping $\downarrow(s - \{X\})$: $\vartheta_s \rightarrow \vartheta_{s - \{X\}}$, called *marginalization* to $s - \{X\}$, such that if $\sigma$ is a valuation for $s$, then $\sigma^{\downarrow(s - \{X\})}$ is a valuation for $s - \{X\}$. We call $\sigma^{\downarrow(s - \{X\})}$ the *marginal* of $\sigma$ for $s - \{X\}$.

## 2.4 COMBINATION

We assume there is a mapping $\otimes: \vartheta \times \vartheta \rightarrow \vartheta$, called *combination*, such that if $\rho$ and $\sigma$ are valuations for $r$ and $s$, respectively, then $\rho \otimes \sigma$ is a valuation for $r \cup s$.

In summary, a *valuation network* consists of a 5-tuple $\{\Psi, \{\Omega_X\}_{X \in \Psi}, \{\tau_1, ..., \tau_m\}, \downarrow, \otimes\}$ where $\Psi$ is a set of variable, $\{\Omega_X\}_{X \in \Psi}$ is a collection of state spaces, $\{\tau_1, ..., \tau_m\}$ is a collection of valuations, $\downarrow$ is the marginalization operator, and $\otimes$ is the combination operator.

## 2.5 MAKING INFERENCE IN VN

In a VN, the combination of all valuations is called the *joint valuation*. Given a VN, we make inferences by computing the marginal of the joint valuation for each variable of interest. If the marginalization and combination operations satisfy some axioms [Shenoy and Shafer 1990], then we can compute the marginals of the joint valuation locally using the Shenoy-Shafer architecture. This is described in the next section.

# 3 COMPUTING MULTIPLE MARGINALS

In this section, we briefly describe the Shenoy-Shafer architecture [Shenoy and Shafer 1990] for computing multiple marginals of the joint valuation using local computation.

In the Shenoy-Shafer architecture, first we construct a join tree, and then we propagate the valuations in the join tree.

## 3.1 JOIN TREES

A *join tree* is a tree whose nodes are subsets of $\Psi$ such that if a variable is in two distinct nodes, then it is in every node on the path between the two nodes [Maier

1983]. The construction of a join tree from a VN is described in [Shenoy 1991, Lauritzen and Shenoy 1996]. Join trees are useful data structures to cache computation.

## 3.2 PROPAGATION IN JOIN TREES

Once we have a join tree, we associate each valuation with a node in the join tree and we propagate the valuations using two rules as follows.

### 3.2.1 Rule 1 (Messages)

Each node sends a message to each of its neighbors. Suppose $\mu^{r \rightarrow s}$ denotes the message from $r$ to $s$, suppose $N(r)$ denotes the neighbors of $r$ in the join tree, and suppose the valuation associated with node $r$ is denoted by $\alpha_r$, then the message from node $r$ to its neighboring node $s$ is given as follows:

$$\mu^{r \rightarrow s} = (\otimes \{\mu^{t \rightarrow r} \mid t \in (N(r) - \{s\})\} \otimes \alpha_r)^{\downarrow r \cap s}$$

In words, the message that $r$ send to its neighbor $s$ is the combination of all messages that $r$ receives from its other neighbors together with its own valuation suitably marginalized. Regarding timing, it is clear that node $r$ sends a message to neighbor $s$ only when $r$ has received a message from each of its other neighbors. A leaf of the join tree has only one neighbor, and therefore it can send a message to its neighbor right away without waiting for any messages.

### 3.2.2 Rule 2 (Marginals)

When a node $r$ has received a message from each of its neighbors, it combines all messages together with its own valuation and reports the results as its marginal. If $\varphi$ denotes the joint valuation, then

$$\varphi^{\downarrow r} = \otimes \{\mu^{t \rightarrow r} \mid t \in N(r)\} \otimes \alpha_r$$

Using Rules 1 and 2, we can compute the marginal of the joint for each subset in the join tree.

Rules 1 and 2 suggest an architecture shown in Figure 1. Each node in the join tree would have two storage registers, one for the input valuation, and one for reporting the marginal of the joint. Also, each edge in the join tree would have two storage registers for the two messages, one in each direction.

# 4 BINARY JOIN TREES

In this section, we introduce the concept of a binary join tree.

A *binary join tree* is a join tree such that no node has more than three neighbors. To explain the importance of a binary join tree, we will describe by means of an example, the inefficiencies of computation in a non-binary join tree.

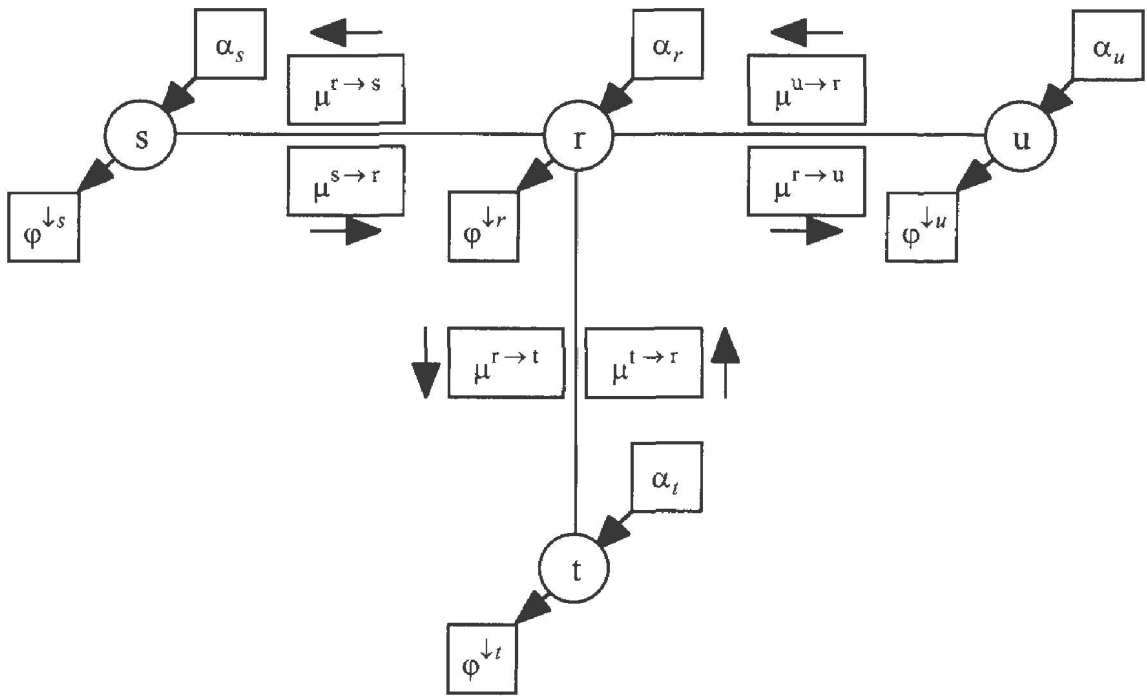

Figure 1: An Architecture for Computing Multiple Marginals

## 4.1 EXAMPLE 1

Consider a valuation network consisting of four variables W, X, Y, and Z, and four valuations $\alpha$ for $\{W, X\}$, $\beta$ for $\{W, Y\}$, $\gamma$ for $\{W, Z\}$, and $\delta$ for $\{X, Y, Z\}$. A non-binary join tree with the messages between adjacent nodes is shown in Figure 2. We make some observations about inefficiencies of computation in this non-binary join tree.

### 4.1.1 Domain of Combination

First, consider the message $(\alpha \otimes \beta \otimes \gamma)^{\downarrow \{X, Y, Z\}}$ (from $\{W, X, Y, Z\}$ to $\{X, Y, Z\}$). The computation of this message involves combination of the valuations $\alpha$, $\beta$, and $\gamma$ on the domain $\{W, X, Y, Z\}$. In general, combination of m valuations on a domain with n configurations involves computation that is linear in $m-1$ and a monotonic increasing function of n. Suppose that W has 2 states, X has 3 states, and Y has 4 states and Z has 5 states. Then the state space of $\{W, X, Y, Z\}$ has 120 configurations. Instead of combining $\alpha$, $\beta$, and $\gamma$ on the domain $\{W, X, Y, Z\}$ that has 120 configurations, it is more efficient to first combine $\alpha$ and $\beta$ on domain $\{W, X, Y\}$ with 24 configurations, and next combine $\alpha \otimes \beta$ with $\gamma$ on the domain $\{W, X, Y, Z\}$ with 120 configurations. A similar observation can be made for the message $(\alpha \otimes \beta \otimes \delta)^{\downarrow \{W, Z\}}$.

### 4.1.2 Non-Local Combination

Second, consider the message $(\beta \otimes \gamma \otimes \delta)^{\downarrow \{W, X\}}$. Notice that Z is in the domain of $\gamma$ and $\delta$, but not in the domain of $\beta$. Thus it follows from one of the axioms that $(\beta \otimes \gamma \otimes \delta)^{\downarrow \{W, X\}} = (\beta \otimes (\gamma \otimes \delta)^{\downarrow \{W, X, Y\}})^{\downarrow \{W, X\}}$. It is computationally more efficient to compute $(\beta \otimes (\gamma \otimes \delta)^{\downarrow \{W, X, Y\}})^{\downarrow \{W, X\}}$ than to compute $(\beta \otimes \gamma \otimes \delta)^{\downarrow \{W, X\}}$. Similarly, instead of computing $(\alpha \otimes \gamma \otimes \delta)^{\downarrow \{W, Y\}}$, it is more efficient to compute instead $(\alpha \otimes (\gamma \otimes \delta)^{\downarrow \{W, X, Y\}})^{\downarrow \{W, Y\}}$.

### 4.1.3 Repetition of Combinations

Third, consider the messages $(\alpha \otimes \beta \otimes \gamma)^{\downarrow \{X, Y, Z\}}$ and $(\alpha \otimes \beta \otimes \delta)^{\downarrow \{W, Z\}}$. Notice that if these two messages are computed separately, then the combination of $\alpha$ and $\beta$ is repeated. Also for messages $(\beta \otimes \gamma \otimes \delta)^{\downarrow \{W, X\}}$ and $(\alpha \otimes \gamma \otimes \delta)^{\downarrow \{W, Y\}}$, the combination of $\gamma$ and $\delta$ is repeated [Xu 1991, Xu and Kennes 1994].

Now consider a binary join tree for the same VN as shown in Figure 3. Compared to the non-binary join tree of Figure 2, the binary join tree has an additional node $\{W, X, Y\}$ and an additional edge $\{\{W, X, Y\}, \{W, X, Y, Z\}\}$.

First, notice that $\alpha \otimes \beta$ is computed on the domain $\{W, X, Y\}$ (as a message from $\{W, X, Y\}$ to $\{W, X, Y, Z\}$) before we compute $(\alpha \otimes \beta \otimes \gamma)^{\downarrow \{X, Y, Z\}}$ (as a message from $\{W, X, Y, Z\}$ to $\{X, Y, Z\}$) and $(\alpha\otimes\beta\otimes\delta)^{\downarrow\{W, Z\}}$ (as a message from $\{W, X, Y, Z\}$ to $\{W, Z\}$). Thus we avoid combining valuations on domains bigger than is necessary.

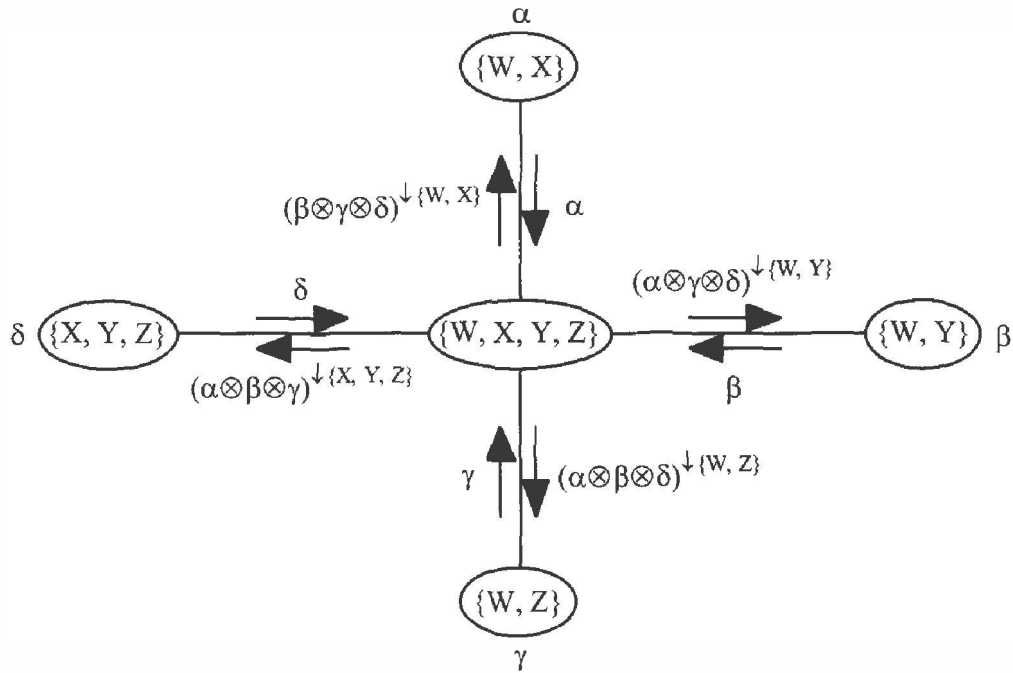

Figure 2: A Non-Binary Join Tree for the VN in Example 1

Second, instead of computing $(\beta\otimes\gamma\otimes\delta)^{\downarrow\{W,X\}}$, we compute $(\beta\otimes(\gamma\otimes\delta)^{\downarrow\{W, X, Y\}})^{\downarrow\{W, X\}}$, and instead of computing $(\alpha\otimes\gamma\otimes\delta)^{\downarrow\{W, Y\}}$, we compute $(\alpha\otimes(\gamma\otimes\delta)^{\downarrow\{W, X, Y\}})^{\downarrow\{W, Y\}}$. Thus the messages are computed locally.

Third, the combination $(\gamma\otimes\delta)^{\downarrow\{W, X, Y\}}$ that appears in messages $(\beta\otimes(\gamma\otimes\delta)^{\downarrow\{W, X, Y\}})^{\downarrow\{W, X\}}$ and $(\alpha\otimes(\gamma\otimes\delta)^{\downarrow\{W, X, Y\}})^{\downarrow\{W, Y\}}$ is computed only once. Also, the combination $\alpha\otimes\beta$ is computed only once for the messages $(\alpha\otimes\beta\otimes\gamma)^{\downarrow\{X, Y, Z\}}$ and $(\alpha\otimes\beta\otimes\delta)^{\downarrow\{W, Z\}}$. Thus we avoid repetition of combinations.

For these three reasons, binary join trees are a more efficient way to organize the computations than non-binary join trees.

How does one construct a binary join tree? We will describe a technique based on the idea of binary combination.

## 4.2 JOIN TREE CONSTRUCTION USING BINARY COMBINATION

We need to structure the join tree so that we combine valuations two at a time. The following procedure does not guarantee a binary join tree. However, it attempts to reduce the number of neighbors of a node.

Let $\Psi$ denote the set of variables, let $\Phi$ denote the set of subsets of variables for which we have valuations or marginals, let the subsets for which we need marginals, let N denote the nodes of the binary join tree, and E denote the edges of the binary join tree, let $|\Phi|$ denote the number of elements of set $\Phi$, and let $\| s \|$ denote the number of elements of the state space of subset $s$. A procedure in pseudocode for constructing a join tree $(N, E)$ using binary combination is as follows.

### 4.2.1 Procedure

INPUT: $\Psi$, $\Phi$
OUTPUT: N, E
**INITIALIZATION**
$\Psi_u \leftarrow \Psi$ {$\Psi_u$ denotes the set of variables in $\Psi$ that have not yet been deleted}
$\Phi_u \leftarrow \Phi$ {$\Phi_u$ denotes the subsets in $\Phi$ that have not yet been arranged in the join tree}
N $\leftarrow \varnothing$
E $\leftarrow \varnothing$
**DO WHILE** $|\Phi_u| > 1$
  Pick a variable $Y \in \Psi_u$
  $\Phi_Y \leftarrow \{s_i \in \Phi_u \mid Y \in s_i\}$.
  **DO WHILE** $|\Phi_Y| > 1$
  $s_1 \leftarrow s_i$ and $s_2 \leftarrow s_j$ where $s_i, s_j \in \Phi_Y$ and $\|s_i \cup s_j\| \le \|s_p \cup s_q\|$ for all $s_p, s_q \in \Phi_Y$
  **IF** $s_1 \subseteq s_2$, **THEN**
  N $\leftarrow$ N $\cup$ $\{s_1, s_2\}$
  E $\leftarrow$ E $\cup$ $\{\{s_1, s_2\}\}$

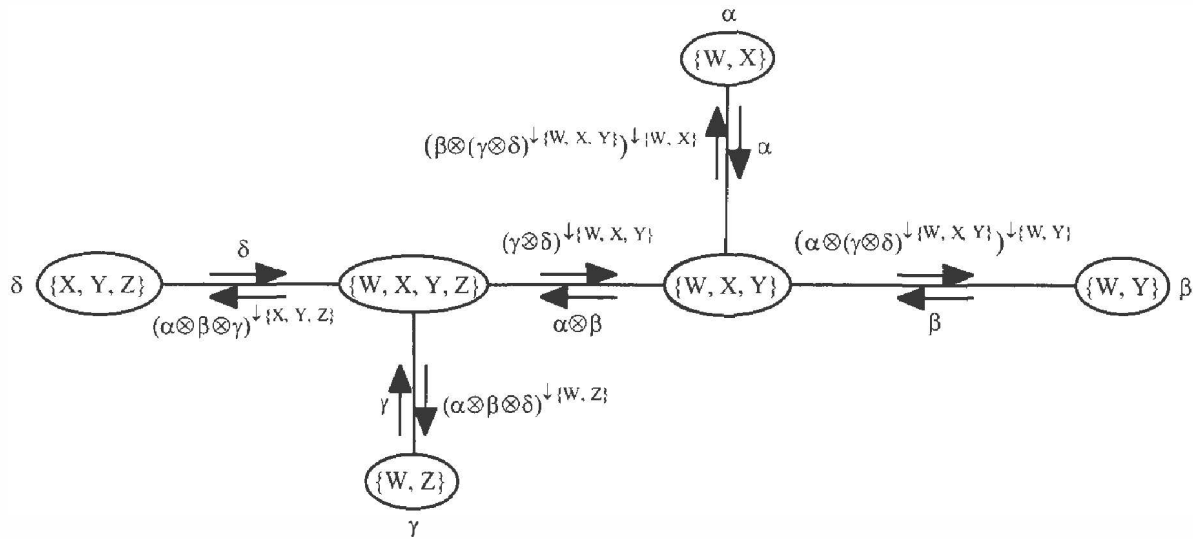

Figure 3: A Binary Join Tree for the VN in Example 1

$\Phi_Y \leftarrow \Phi_Y - \{s_1\}$
**ELSE**
$N \leftarrow N \cup \{s_1, s_2, s_1 \cup s_2\}$
$E \leftarrow E \cup \{\{s_1, s_1 \cup s_2\}, \{s_2, s_1 \cup s_2\}\}$
$\Phi_Y \leftarrow \Phi_Y - \{s_1, s_2\} \cup \{s_1 \cup s_2\}$
**END  IF**
**END  DO**
$s \leftarrow s_i$ where $\{s_i\} = \Phi_Y$
$N \leftarrow N \cup \{s\} \cup \{s - \{Y\}\}$
$E \leftarrow E \cup \{\{s, s - \{Y\}\}\}$
$\Phi_u \leftarrow \Phi_u \cup \{s - \{Y\}\} - \{s_i \in \Phi_u \mid Y \in s_i\}$
**END  DO**
$N \leftarrow N \cup \Phi_u$
**END**

### 4.3  EXAMPLE 2

Consider a valuation network with valuations as follows:
$\delta$ for $\{D\}$, $\sigma_1$ for $\{D, S_1\}$, $\sigma_2$ for $\{D, S_2\}$, $\sigma_3$ for $\{D, S_3\}$,
$\sigma_4$ for $\{D, S_4\}$, $o_1$ for $\{S_1\}$, and $o_2$ for $\{S_2\}$. Suppose we
need the marginal of the joint for all five variables. If we
implement the binary combination procedure for the
collection $\Phi = \{\{D\}, \{D, S_1\}, \{D, S_2\}, \{D, S_3\}, \{D, S_4\},$
$\{S_1\}, \{S_2\}, \{S_3\}, \{S_4\}\}$ using deletion sequence
$S_1 S_2 S_3 S_4$, the resulting join tree displayed in Figure 4 is
non-binary.

Figure 4 also displays the messages computed using
Rules 1 and 2 for the marginals of each variable in the
VN. Notice that although the join tree is non-binary, the
computation of the messages does not exhibit the
inefficiencies labeled domain of combination and non-local
combination exhibited by Example 1. However, notice
that the messages from node $\{D\}$ to its four neighbors
does exhibit the inefficiency labeled repetition of
combinations. For example, the message from $\{D\}$ to $\{D,$

$S_1\}$ and the message from $\{D\}$ to $\{D, S_2\}$ share the
combination $\sigma_3^{\downarrow D} \otimes \sigma_4^{\downarrow D} \otimes \delta$. The processor at node $\{D\}$
does 13 combinations in total (3 in each of the 4
messages as shown in Figure 4, and 1 more to compute
the marginal for $\{D\}$ not shown in Figure 4).

One way to avoid the repetition of combinations is for the
processor at $\{D\}$ to combine valuations two at a time and
to cache the intermediate results. A more explicit way to
avoid the repetition of combinations is to make the join
tree binary as shown in Figure 5. We create multiple
copies of node $\{D\}$ and connect them together as shown in
Figure 5. The input valuation $\delta$ is associated with only
one $\{D\}$ node, the $\{D\}$ node connected to $\{D, S_4\}$. The
$\{D\}$ node connected to $\{D, S_1\}$ and $\{D, S_2\}$ does the
combination $\sigma' \otimes \sigma''$ where $\sigma' = (\sigma_1 \otimes o_1)^{\downarrow D}$ and $\sigma'' =$
$(\sigma_2 \otimes o_2)^{\downarrow D}$, the $\{D\}$ node connected to $\{D, S_3\}$ does the
combination $\sigma''' \otimes \sigma_3$ where $\sigma''' = \sigma' \otimes \sigma''$, and the $\{D\}$
node connected to $\{D, S_4\}$ does the combination $\sigma'''' \otimes \delta$
where $\sigma'''' = \sigma''' \otimes \sigma_3$.

A sketch of the procedure for constructing the binary join
tree shown in Figure 5 is as follows. We start with the
non-binary join tree as shown in Figure 4 obtained by the
binary combination procedure. Next, we designate a node
as the root, say $\{S_4\}$, and direct the edges toward the root.
Next, we write down the messages computed by each node
for the computation of the marginal for the root. If a node
needs to do more than one combination, we create
multiple copies of the node so that only one binary
combination is necessary at each node. For example, in
the join tree of Figure 4, the message from $\{D\}$ to $\{D,$
$S_4\}$ is $(\sigma_1 \otimes o_1)^{\downarrow D} \otimes (\sigma_2 \otimes o_2)^{\downarrow D} \otimes \sigma_3^{\downarrow D} \otimes \delta$ which consists
of three combination operations. If we perform the three
binary combination at three different $\{D\}$ nodes as
suggested by $((((\sigma_1 \otimes o_1)^{\downarrow D} \otimes (\sigma_2 \otimes o_2)^{\downarrow D}) \otimes \sigma_3^{\downarrow D}) \otimes \delta)$,
then we obtain the binary join tree structure shown in
Figure 5.

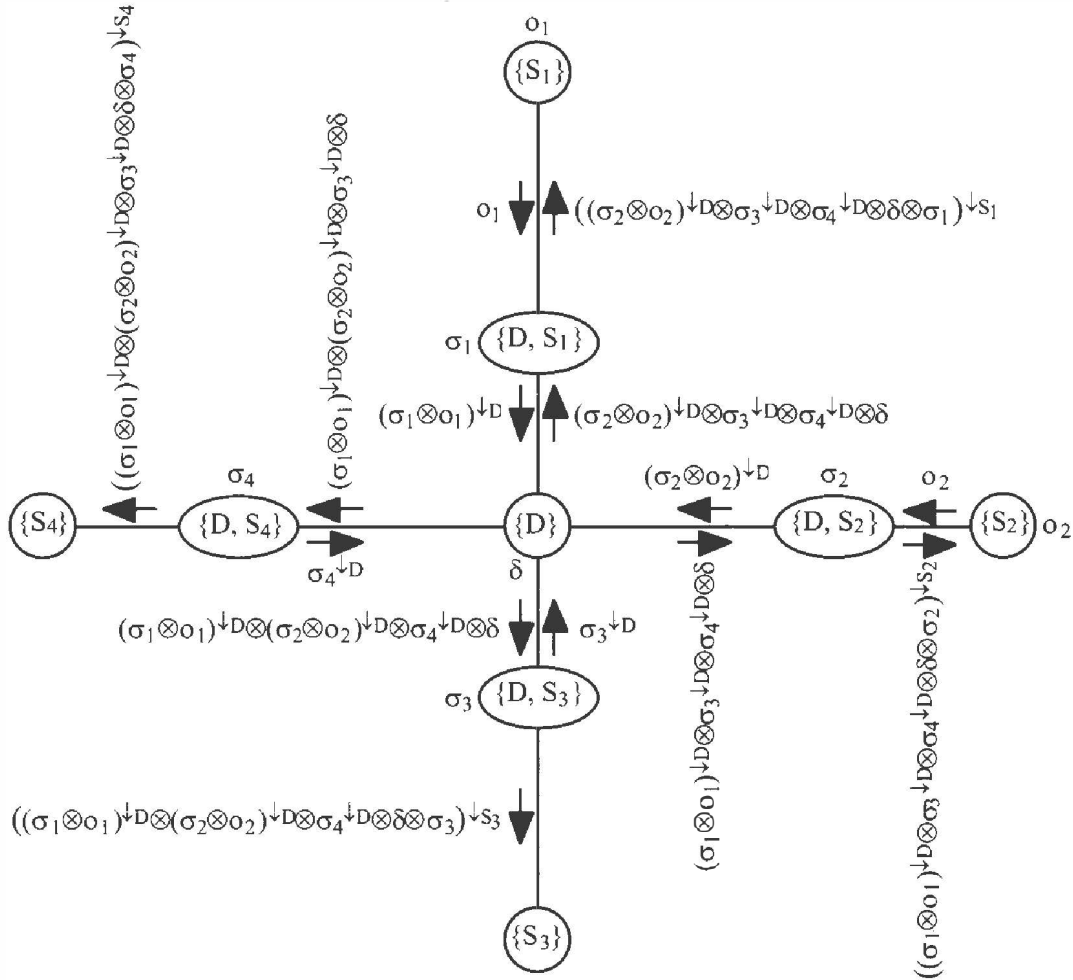

Figure 4: A non-binary join tree for the VN in Example 2

Notice that the computations of messages in the binary join tree do not suffer from the repetition of combinations phenomenon. The processors at the three {D} nodes do a total of 9 combinations (3 by the {D} node connected to {D, $S_1$}, 3 by the {D} node connected to {D, $S_3$}, and 3 by the {D} node connected to {D, $S_4$} assuming that this {D} node computes the marginal for {D}). The messages are shown in Figure 5, but the computation of the marginals are not shown.

## 5  CONCLUSION

The main goal of this paper is to describe a data structure called binary join trees that are useful in computing multiple marginals efficiently using local computation. We define binary join trees, describe their utility, and sketch a procedure for constructing them.

The join tree construction process described here is superficially different from the method described in

Lauritzen and Spiegelhalter [1988] which consists of moralizing a directed acyclic graph, triangulating the moral graph using the maximum cardinality search method, and then arranging the cliques of the triangulated moral graph in a join tree. Instead of starting from a directed acyclic graph, we start with a more general setting—a hypergraph consisting of all subsets for which we have valuations (this is roughly the same as the cliques of a moral graph) and all subsets for which we desire marginals, and then we use the fusion algorithm [Shenoy 1992, Cannings *et al.* 1978] as a guide for constructing a join tree. Alternative procedures for join tree construction have been suggested by Draper [1995].

A join tree can be regarded as a data structure to organize the computations involved in computing multiple marginals. Binary join trees further refine the data structure so that unnecessary computations are minimized. In particular, we identify three sources of inefficiencies

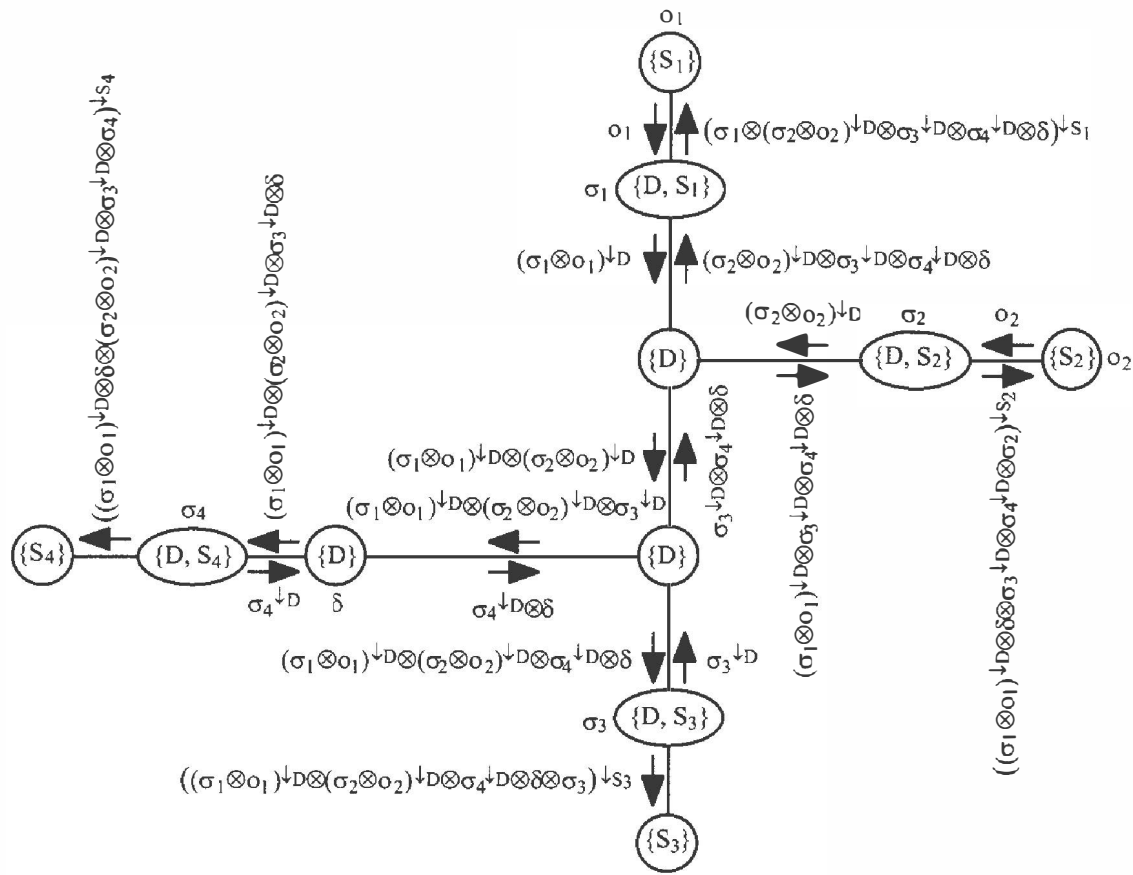

Figure 5: A Binary Join Tree for the VN in Example 2

associated with non-binary join trees that are eliminated in binary join trees.

A complete version of this paper [Shenoy 1995] is available via anonymous ftp from the author's www homepage.

### Acknowledgments

This work is based on research supported in part by Hughes Research Laboratories under grant No. GP3044.94-075. I am grateful to Liping Liu, Hong Xu, Steffen L. Lauritzen, Finn V. Jensen, Peter Gillett, Yang Chen, and anonymous referees for comments and discussions.